\documentclass{article}

\usepackage[a-1b]{pdfx}

    \PassOptionsToPackage{numbers, sort, compress}{natbib}

\usepackage[sglblindworkshop, final]{neurips_2025}
\workshoptitle{ML for Systems}
 
\usepackage{natbib}
\usepackage{amsmath,amssymb,amsfonts}
\usepackage{algorithmic}
\usepackage{graphicx}
\usepackage{textcomp}
\usepackage{xcolor}

\usepackage{subcaption}
\usepackage{booktabs}
\usepackage{titlesec}

\def\BibTeX{{\rm B\kern-.05em{\sc i\kern-.025em b}\kern-.08em
    T\kern-.1667em\lower.7ex\hbox{E}\kern-.125emX}}

\newcommand{\ds}[1]{\textbf{\color{orange}[DS: #1]}}

\titlespacing*\section{0pt}{0pt plus 4pt minus 2pt}{0pt plus 2pt minus 2pt}
\titlespacing*\subsection{0pt}{0pt plus 2pt minus 2pt}{0pt plus 2pt minus 2pt}
\titlespacing*\subsubsection{0pt}{0pt plus 2pt minus 2pt}{0pt plus 2pt minus 2pt}

\title{A Joint Learning Approach to Hardware Caching and Prefetching}
\author{%
  \textbf{Samuel Yuan}$^{1}$ \quad
  \textbf{Divyanshu Saxena}$^{1}$ \quad
  \textbf{Jiayi Chen}$^{1}$ \quad
  \textbf{Nihal Sharma}$^{1,2 *}$ \quad
  \textbf{Aditya Akella}$^{1}$\\[0.25em]
  $^{1}$The University of Texas at Austin \quad $^{2}$Capital One\\[0.25em]
  {\tt \{syuan, janechen, nihal.sharma\}@utexas.edu  \hfill \{dsaxena, akella\}@cs.utexas.edu}
}

\begin{document}
\maketitle
\def\thefootnote{*}\footnotetext{This work was conducted when Nihal Sharma was at  The University of Texas at Austin.}\def\thefootnote{\arabic{footnote}}


\begin{abstract}

Several learned policies have been proposed to replace heuristics for scheduling, caching, and other system components in modern systems.
By leveraging diverse features, learning from historical trends, and predicting future behaviors, such models promise to keep pace with ever-increasing workload dynamism and continuous hardware evolution.
However, policies trained in isolation may still achieve suboptimal performance when placed together.
In this paper, we inspect one such instance in the domain of hardware caching -- for the policies of cache replacement and prefetching.
We argue that these two policies are bidirectionally interdependent and make the case for training the two \textit{jointly}. 
We propose a joint learning approach based on developing shared representations for the features used by the two policies.
We present two approaches to develop these shared representations, one based on a joint encoder and another based on contrastive learning of the embeddings, and demonstrate promising preliminary results for both of these.
Finally, we lay down an agenda for future research in this direction.

\end{abstract}

\section{Introduction}



Hardware cache management has been one area of systems research where machine learning (ML) based methods have been shown to outperform manual heuristics, particularly in prefetching~\cite{zhan18-voyager, bhatia19, hashemi18, bera2021pythia, duong2024new}, and cache replacement~\cite{zhan19cr-glider, jimenez17, jaleel2010high}.
Prefetching aims to reduce memory access latency and improve throughput by predicting and loading data into the cache before it is explicitly requested.
Cache replacement policies, in contrast, determine which blocks to evict to make space for new ones, directly influencing hit rates and overall efficiency.

Prefetching and cache replacement policies are highly interdependent on each other.
An effective cache replacement strategy can use the knowledge of future prefetches as a factor in its eviction decisions, and a cache-aware prefetcher can refrain from fetching data that is already likely to remain resident in cache. 
Conversely, `incompatible' policies can cause performance degradation -- an aggressive prefetcher can flood the cache with objects that are immediately evicted, while a replacement policy may evict prefetched blocks just before they are used.

Although both prefetching and replacement have been studied extensively, prior work on learning-based approaches has largely treated the two as separate problems.
Learned cache replacement algorithms either do not account for the presence of a prefetcher in their design~\cite{jain16-hawkeye, zhan19cr-glider, reuse-perceptron-micro16, reuse-prediction-micro17}, and even when they are designed in a prefetcher-aware manner~\cite{shah22-mockingjay}, the prefetcher is typically a fixed, non-learned one.
Similarly, learned prefetchers are usually trained on traces collected under default cache replacement policies~\cite{zhan18-voyager, rl-prefetcher-isca15, neural-prefetcher-taco19}.
As we argue in Section~\ref{sec:motivation}, this separation is problematic.
As is the case with learning algorithms, learned caching and prefetching policies trained in isolation implicitly assume that the policies will only observe inputs from the training distribution.
When exposed to inputs outside the training distribution, the policies can result in sub-optimal behavior.
A special case of this sub-optimality occurs when policies for prefetching and caching, trained in isolation, are placed together. 
In this paper, we attempt to answer the question of whether there is merit in a joint learning regime for caching and prefetching.

Our proposal is to use shared representations for the input features of the two algorithms to capture the interactions between them.
At a high level, this shared representation is obtained by an encoder that maps the input features into an embedding space that captures the relationships between features of the two policies.
The resulting embeddings are then provided to both models, enabling each to be aware of the other’s features during training.

In this paper, we present two approaches for developing these shared representations.
The first approach is based on \textbf{joint encoding}, where the features used by prefetching and replacement algorithms are fed into a shared encoder.
When this embedding is used by the downstream policy, it can learn useful correlations between the two feature sets from this shared representation.
The second approach relies on \textbf{contrastive learning}~\cite{lekhac20, he20, altahan20, chen20, pmlr-v119-laskin20a}, where the encoder is trained by contrasting feature pairs that are temporally co-occurring as positive pairs against feature pairs that do not occur together as negative pairs.
This encourages the encoder to align the embeddings of feature sets that occur together, thereby capturing the correlations between the features.
While the first approach implicitly allows the model to discover interactions, contrastive learning enforces alignment directly.
More details on each approach and their pros and cons are discussed in Section~\ref{sec:approach}.

We apply these two approaches to existing state-of-the-art data prefetching and replacement models.
Crucially, we retain the features, outputs and training data for both of these models.
Our preliminary results demonstrate that by simply using a more informed joint training regime, the performance accuracy of models significantly improves.
Taking a broader view, the techniques described in this paper can also be applied to other domains -- for example, independently tuned congestion-control and flow-scheduling policies, or separately learned scheduling and admission-control policies, where incoherent policies may degrade one another’s effectiveness.
\section{Background and Motivation}
\label{sec:motivation}



Both cache replacement and prefetching share the same fundamental goal: understanding cache access patterns in order to maximize effective cache utilization.

Traditional learned cache replacement policies are trained to learn reuse likelihood from features such as instruction PCs, memory addresses, block offsets, or recency/frequency history, and use these signals to decide which cachelines to evict. For example, Hawkeye~\cite{jain16-hawkeye} leverages a PC-based binary classifier that labels loads as cache-friendly or cache-averse and inform eviction decisions accordingly. 

Learned prefetchers attempt to identify recurring access patterns. They are often trained on correlations between program context and observed address streams to issue predictions for addresses that are likely to be needed in the near future. For example, Voyager~\cite{zhan18-voyager} learns correlations between PCs and subsequent addresses to generate targeted prefetch requests.

Although both policies rely on understanding access patterns, prior works have largely studied replacement and prefetching in isolation. Replacement assumes that every line was demand-fetched and therefore encodes meaningful reuse information, while prefetching assumes that every line it fetches will be judged fairly by the replacement policy. In practice, however, these two mechanisms are bidirectionally interdependent: If the prefetcher brings in a line that the replacement policy classifies as cache-averse and immediately evicts, the prefetch is wasted; on the other hand, if the replacement predictor choose to keep the prefetched lines, they could pollute the cache. Training the two policies separately therefore leads to suboptimal behavior.

Mockingjay~\cite{shah22-mockingjay} makes an attempt on prefetching-aware replacement by preferentially evict lines that will be prefetched in the future.
It has two key limitations:
(i) The coupling studied by Mockingjay is one-way: the prefetcher is not replacement-aware, so it may still make wasteful requests that consume bandwidth and cache space.
(ii) Mockingjay assumes the prefetcher’s decisions are already high quality, but in practice prefetch accuracy varies widely across workloads. For each new workload and new prefetcher, Mockingjay requires training from scratch.

These limitations highlight a deeper insight: replacement and prefetching draw from the same underlying feature space, and their predictions are fundamentally entangled. Rather than training two separate models that interact only indirectly, a more effective solution is to learn them jointly.

\section{Joint Learning}
\label{sec:approach}


The key challenge in performing joint learning is in designing training mechanisms that make the models aware of the other policy's decisions.
Taking inspiration from prior work on joint policies for multi-agent systems~\cite{maddpg-neurips17,qmix-pmlr-rashid18a,commnet-neurips16}, we propose developing shared representations for these interdependent policies.
Such representations allow one model’s decisions to be informed by the other’s context, thereby facilitating coordination.
For concreteness, we ground our study in the Glider cache replacement policy~\cite{zhan19cr-glider} and the Voyager prefetching policy~\cite{zhan18-voyager}.

\begin{figure*}[t]
    \centering
    \begin{subfigure}[b]{0.29\textwidth}
        \centering
        \includegraphics[width=\textwidth]{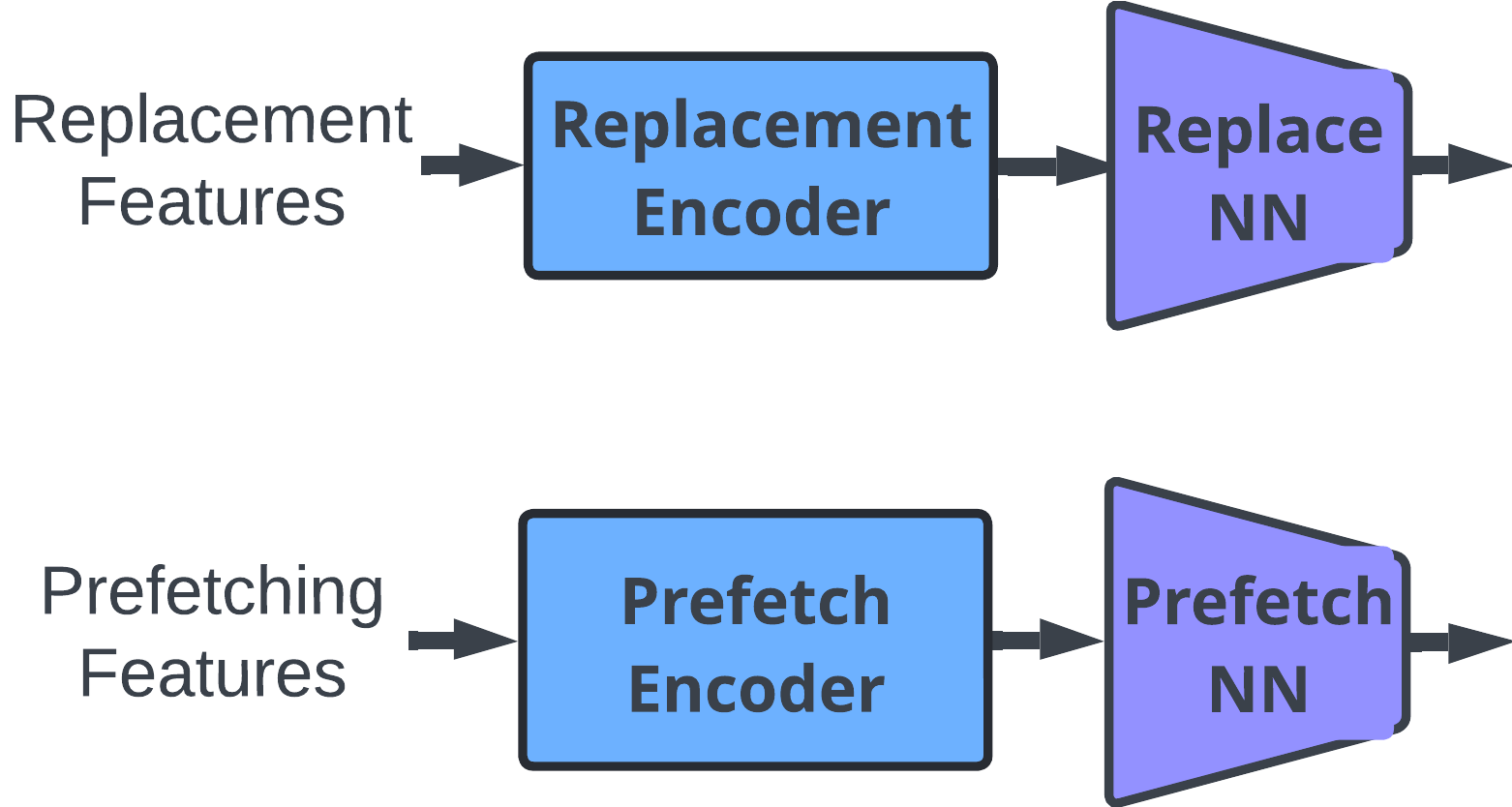}
        \caption{\small Baseline models.}
        \label{fig:baseline-models}
    \end{subfigure}
    \hfill
    \begin{subfigure}[b]{0.38\textwidth}
        \centering
        \includegraphics[width=\textwidth]{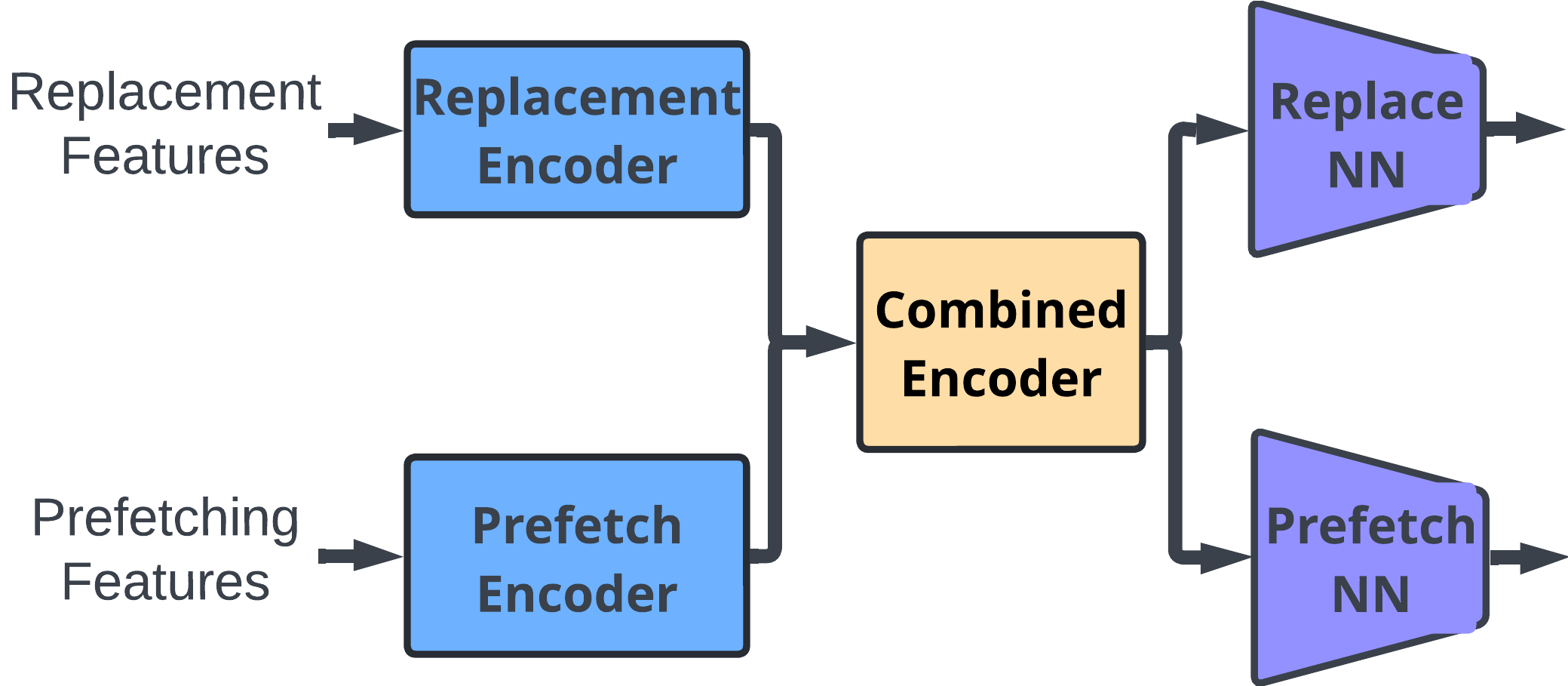}
        \caption{\small Training using joint encoder.}
        \label{fig:encoder-models}
    \end{subfigure}
    \hfill
    \begin{subfigure}[b]{0.29\textwidth}
        \centering
        \includegraphics[width=\textwidth]{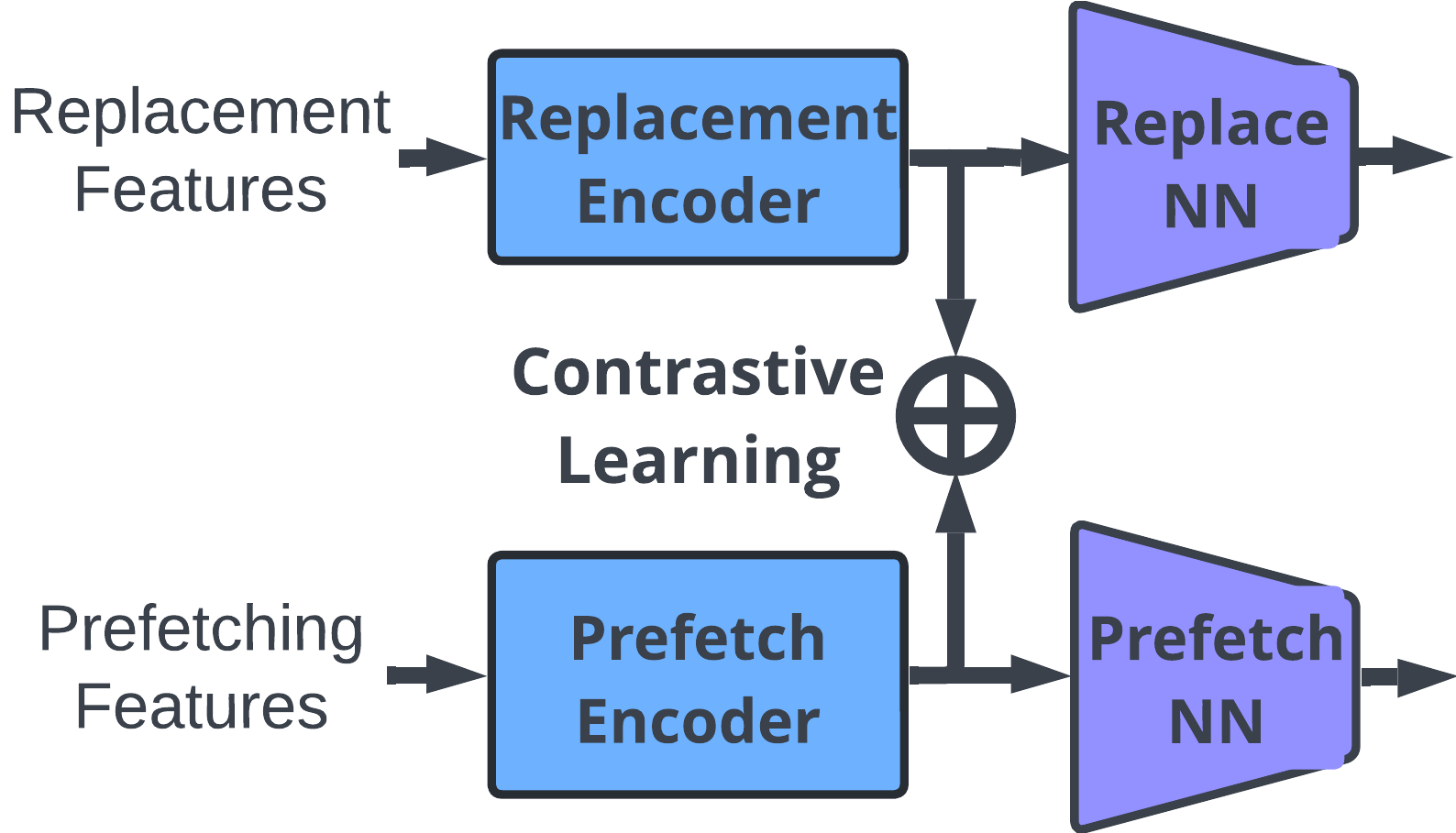}
        \caption{\small Contrastive Learning.}
        \label{fig:contrastive-models}
    \end{subfigure}

    \caption{Logical diagrams for the model architectures for state-of-the-art learned replacement and prefetching policies (a), versus our proposed joint learned models: (b) using a joint encoder, and (c) using contrastive learning.}
    \label{fig:models}
\end{figure*}

The architectures used by these models can be largely broken into two steps (see Figure~\ref{fig:baseline-models}): an encoding step, where the input features (addresses and instruction pointers) are translated into useful representations; and a learning step, where the representations are used to learn policy decisions.
For the Glider replacement policy, the encoder is an embedding layer that produces an embedding from a sequence of program counters (PC).
This embedding is then fed into a simple 2-layer LSTM-based model.
Similarly, for the Voyager prefetching policy, a set of embedding layers encode the input features (which includes the PC, page, and offset sequences) followed by another set of LSTM-based prediction layers.
The key observation here is that the embedding layers in both of these models serve to translate the large input spaces of addresses and PC counters into an embedding that allows neural architectures to learn easily~\cite{zhan18-voyager}.

In our approach, we repurpose these embedding layers to not just translate the input spaces into an embedding, but also learn embeddings that capture the features observed by the model of the other interdependent policy.
In this work, we explore two ways of developing shared representations.

\paragraph{Shared representations via joint encoder.}
The first approach is to augment the individual encoders with a combined encoder that learns a joint embedding space across replacement and prefetching features (see Figure~\ref{fig:encoder-models}).
During training, the outputs of the individual encoders are passed through the combined encoder, and all encoders and downstream neural networks are trained together. Backpropagation flows from the loss through the combined network into the concatenated embeddings, splitting the gradient so that each half propagates into its respective encoder (cache and prefetch). Thus, all encoders and the joint head are updated end-to-end by the same objective.

Training the combined encoder jointly with the individual encoders and policy networks ensures that the embeddings evolve to reflect shared structural information.

\paragraph{Aligning representations via contrastive learning.}
The second approach leverages contrastive learning to align the representations learned by the two encoders (see Figure~\ref{fig:contrastive-models}).
Here, the training proceeds in two steps.
First, the individual encoders are trained against a contrastive loss, such that embeddings of temporally correlated features from replacement and prefetching are treated as positive pairs, while uncorrelated pairs are treated as negatives. For example, consider a prefetch decision that brings a block into the cache, as well as a replacement decision that happens shortly thereafter. Because these two events are temporally linked, their embeddings are treated as a positive pair and pulled closer together in the representation space. This contrastive setup pulls embeddings of causally related events together while pushing unrelated activity apart. 
This structured training objective creates alignment between the two encoders without requiring an explicit combined encoder.
Once these encoders are trained, the downstream policy neural networks are also trained.
In this second stage, each policy network is trained using the same objective function as their individual counterparts. 

\paragraph{On the pros and cons of each approach.}
When shared representations are obtained using the joint encoder, the encoders and the policy networks are trained together -- allowing the shared embeddings to evolve along with the networks used for decision-making for downstream tasks.
However, this is an implicit signal that may not work for joint learning approaches in general -- the embedding can entangle features that are irrelevant across tasks.
Moreover, it requires synchronous availability of features from both models, which may not always be practical. 
On the other hand, the contrastive learning approach enforces an explicit alignment signal. 
When the model observes enough data points of positive and negative samples (in our case, features from replacement and prefetching models that occur together versus not), it ensures that temporally correlated features are explicitly brought closer in the embedding space.

\section{Preliminary Results and Future Directions}
\label{sec:impl-eval}

\begin{table}[t]
\centering
\caption{Cache replacement performance comparison for the two joint-learned alternatives against the uncoordinated baseline across workloads. Best results in each column are presented in bold.}
\label{tab:workloads}
\begin{tabular}{lccccc}
\toprule
\textbf{Method} & \textbf{605.mcf} & \textbf{401.bzip2} & \textbf{437.leslie3d} & \textbf{623.xalancbmk} & \textbf{620.omnetpp} \\
\midrule
Cache-only         & 81.14\% & \textbf{88.72\%} & 78.29\% & 92.65\% & 91.97\% \\
Cache-encoder & \textbf{90.33\%} & 88.71\% & 99.67\% & \textbf{97.07\%} & \textbf{92.13\%} \\
Cache-contrastive & 81.39\% & \textbf{88.72\%} & \textbf{99.70\%} & 93.21\% & 91.81\% \\
\bottomrule
\end{tabular}
\end{table}

We implement our approach by extending the ChampSim simulator~\cite{champsim} with data collection, oracle labeling, and model training pipelines.
We test our hypothesis by running the three approaches shown in Figure~\ref{fig:models} over five SPEC CPU traces on ChampSim.

\paragraph{Training Setup}
We use SPEC CPU traces in ChampSim to collect data for training our models.
The cache replacement predictor is trained on the first 60\% of each dataset, validated on the next 20\%, and evaluated on the last 20\%.
We collect features for both replacement and prefetching by logging the state of both components whenever a tag lookup occurs.
Each logged event records the 64-byte block address (aligned physical address) and the program counter (IP) of the demand or prefetch request.
We also record contextual features such as core ID, miss type, stride/stream identifier, set index, and cycle count.
To obtain ground-truth labels, we replay each trace offline and run the Belady’s MIN algorithm.
This produces the optimal replacement decision for every cache miss.
We classify each line insertion as \textit{cache-friendly} if the block receives at least one future hit before eviction and \textit{cache-averse} if the block is never reused (i.e., dead-on-arrival or bypassed).

We use this training regime for the three alternatives shown in Figure~\ref{fig:models}, where each model is trained with the same set of oracle labels and evaluated on held-out SPEC traces.
This setup allows us to directly quantify the impact of joint representation learning and contrastive pretraining relative to conventional cache-only predictors.

\paragraph{Results}
We measure the predictor’s accuracy as the fraction of inputs for which its decision, to label a sequence of PC addresses as cache-friendly or cache-averse, matches the oracle’s decision (Belady’s MIN algorithm in our case).
Note that a high accuracy implies that the replacement model is able to match the decisions of an oracle, directly leading to improved cache hit ratios~\cite{zhan19cr-glider}.
Table~\ref{tab:workloads} shows the obtained accuracies for the three alternatives -- using the joint learning approach results in better performance across all five traces, providing 1-1.3$\times$ better accuracy. 
In particular for the {\em leslie3d} trace, the jointly trained models achieve almost 100\% accuracy.

\paragraph{Future Directions}
Our proposal opens up several avenues of interesting research.
First, translating these techniques into practical hardware requires lightweight models, motivating the use of distillation and quantization techniques to minimize inference overhead.
Second, a deeper investigation into the interpretability of the learned representations could shed light on how coordination between prefetching and replacement emerges in practice.
Another interesting exploration is on applying these techniques to other pairs of interdependent policies (such as congestion control and flow scheduling, or scheduling and admission control), which may surface additional challenges such as decision-making overheads, reliance on stale features, or heterogeneity in learning algorithms—for instance, enabling joint training when one model leverages online reinforcement learning while the other is trained with supervised learning.

\section*{Acknowledgements}

We thank the anonymous reviewers, members of the UTNS lab and the LDOS project for their feedback that helped improve the paper. 
This material is based upon work supported by the U.S. National Science Foundation (NSF) under Grant Number 2326576.

\bibliographystyle{plainnat}
\bibliography{refs}

\end{document}